\documentclass{article}
\usepackage{spconf,amsmath,graphicx}
\usepackage{mathtools}
\usepackage{caption,indentfirst,placeins}
\usepackage{svg}
\usepackage{booktabs}

\renewenvironment{thebibliography}[1]{
  \begin{oldthebibliography}{#1}
    \setlength{\itemsep}{0pt plus 0.3ex}
    \setlength{\parskip}{0pt plus 0.3ex}
}
{
  \end{oldthebibliography}
}

\title{Social Relation Recognition in Egocentric Photostreams}
\name{Emanuel S\'anchez Aimar$^{(1)}$, Petia Radeva$^{(1)}$, Mariella Dimiccoli$^{(2)}$}

\address{%
$^{(1)}$ University of Barcelona, Computer Vision Center, Barcelona, Spain.\\
$^{(2)}$  Institut de Rob\`otica i Inform\`atica Industrial, CSIC-UPC, Barcelona, Spain
}

\begin{document}
%
\maketitle
\begin{abstract}
This paper proposes an approach to automatically categorize the social interactions of a user wearing a photo-camera ($2$fpm), by relying solely on what the camera is seeing. The problem is challenging due to the overwhelming complexity of social life and the extreme intra-class variability of social interactions captured under unconstrained conditions. We adopt the formalization proposed in Bugental's social theory, that groups human relations into five social domains with related categories. Our method is a new deep learning architecture that exploits the hierarchical structure of the label space and relies on a set of social attributes estimated at frame level to provide a semantic representation of social interactions. Experimental results on the new \textit{EgoSocialRelation} dataset demonstrate the effectiveness of our proposal. 

\end{abstract}


\section{Introduction}
\label{sec:intro}

As our social life keeps moving towards the digital world and its social networks, new collective moments are continuously being captured in the form pictures, audio, videos, and text. Meanwhile, several studies have shown that human relationships have an important effect in human health, involving physical and mental health, behaviour, and mortality risk \cite{umberson2010health}. The need of a broader understanding of our social relations and their influence on human health have motivated an increasing interest in the computer vision community for automatic discovery, quantification and categorization of social interactions from the vast amount of public images and videos \cite{Ramanathan2013,dehghan2014look,murillo2012urban,shao2013you,sun2017domain,aghaei2018towards}. Recently, Aghaei \textit{et al.} \cite{aghaei2018towards} have shown the usefulness of  \textit{egocentric photostreams}, captured by a wearable photo-camera \cite{bolanos2017towards} to automatically analyze the daily social interactions of a person, in a natural setting where people appear in an intimate perspective. Despite the challenging characteristics of the egocentric domain, such as the fact that the user is not visible in the field of view, background clutter, and abrupt appearance changes \cite{aghaei2018towards}, the authors showed that it is possible not only to understand when the camera wearer is interacting with somebody, but also to determine with how many people the user has interacted with during a given period of time, the duration and frequency of the interactions. However, in \cite{aghaei2018towards} the classification of interactions was limited to formal and informal meetings. Recent work \cite{sun2017domain} has proposed the Bugental's domain-based social theory \cite{Bugental2000} as a conceptualization of human social life to categorize social interactions in images. The theory includes five domains with examples of common relations, characterized by specifics attributes and behaviours. Nevertheless, in \cite{sun2017domain}, this approach has been applied on  a dataset of third-person images collected from photo albums.

\begin{figure}[t]
     \centering
     \includegraphics[width=0.45\textwidth]{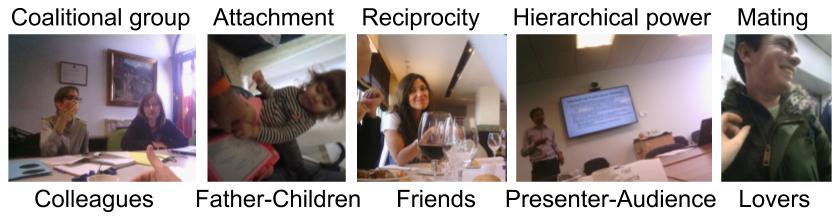}
     \caption{Examples of images from \textit{EgoSocialRelation} dataset.}
     \label{fig:more-egocentric-images}
\end{figure}

To evaluate the formalization proposed in \cite{sun2017domain} in a naturalistic setting and at the same time going deeper into the understanding of social interactions from \textit{egocentric photostreams}, in this work we propose a new egocentric dataset, hereafter referred to as \textit{EgoSocialRelation} dataset\footnote{
https://chest.iri.upc.edu/files/users/mdimiccoli/public\_html/DATASETS/\\EgoSocialRelation.zip}, where social interactions, formed of short image sequences, are annotated in a hierarchy of domain and relation labels, derived from Bugental's theory. Furthermore, we propose and validate, for the first time on egocentric data, several models for social relation categorization, hence providing a solid benchmark for further studies. The proposed models rely on the composition of visual semantic attributes, and exploit the sequential nature of photostreams. Code for experiments is publicly released\footnote{https://github.com/emasa/social-relations-recognition-egocentric-photostreams}.

The rest of the paper is as follows: section \ref{sec:related-work} reviews related work; section \ref{sec:methodology} details our approach; section \ref{sec:results} introduces a new dataset and discusses experimental results. Section \ref{sec:conclusions} concludes this work summarizing its contributions.

\section{Related work}
\label{sec:related-work}
\noindent{\bf Third-view domain\quad} 
A large body of work has focused on family member recognition \cite{dehghan2014look}, role recognition in social events \cite{Ramanathan2013}, categorization of social groups based on appearance (e.g. urban tribes \cite{murillo2012urban}) and occupation recognition based on contextual information \cite{shao2013you}. Inspired on Bugental's social domain-based theory \cite{Bugental2000}, the authors of \cite{sun2017domain} proposed an \textit{holistic} approach to social relations categorization from image data. The published dataset contains \textit{still images} with labels in a hierarchical structure of 5 social domains and 16 social relations, derived from Bugental's theory. Furthermore, they derived a family of models, combining \textit{human attributes} and \textit{behaviours}, extracted with pretrained Convolutional Neural Networks (CNN), to ultimately classify the CNN feature representation with a Support Vector Machine (SVM).\\
\noindent{\bf Egocentric domain\quad} The seminal work of Fathi \textit{et al.} \cite{Fathi2012} proposed to detect social interactions from egocentric videos, by inferring the 3D location to which a person is looking at, through a Markov Random Field model. They further categorized these interactions into three classes, namely \textit{discussion}, \textit{dialogue} and \textit{monologue}, depending on the role of the participants in the interaction. Later on, \cite{yang2016wearable} proposed a procedure to analyze social interaction sequences from egocentric videos and detect them applying a model based on Hidden Markov Model and SVM, focusing on head and body poses. With a different approach, Aghaei \textit{et al.} \cite{aghaei2015towards,aghaei2016whom} proposed to detect social interactions in egocentric photostreams, while exploiting distance and orientation of observable people w.r.t. the camera-user. Extending the previous work, \cite{aghaei2018towards} categorized interactions into two broad categories, namely \textit{formal} and \textit{informal meetings}, based on the classification of sequences with a Long Short Term Memory (LSTM) model \cite{Hochreiter1997lstm}. The authors extended the original attributes with facial expressions and global CNN features extracted at frame level, the former related to affective internal states and the latter to account for contextual information. They argue that for effective detection and categorization of social interactions from an egocentric perspective, a combination of social signals and environmental features is needed, as well as their evolution over time.
\begin{figure}[t]
    \centering
    \includegraphics[width=0.3\textwidth]{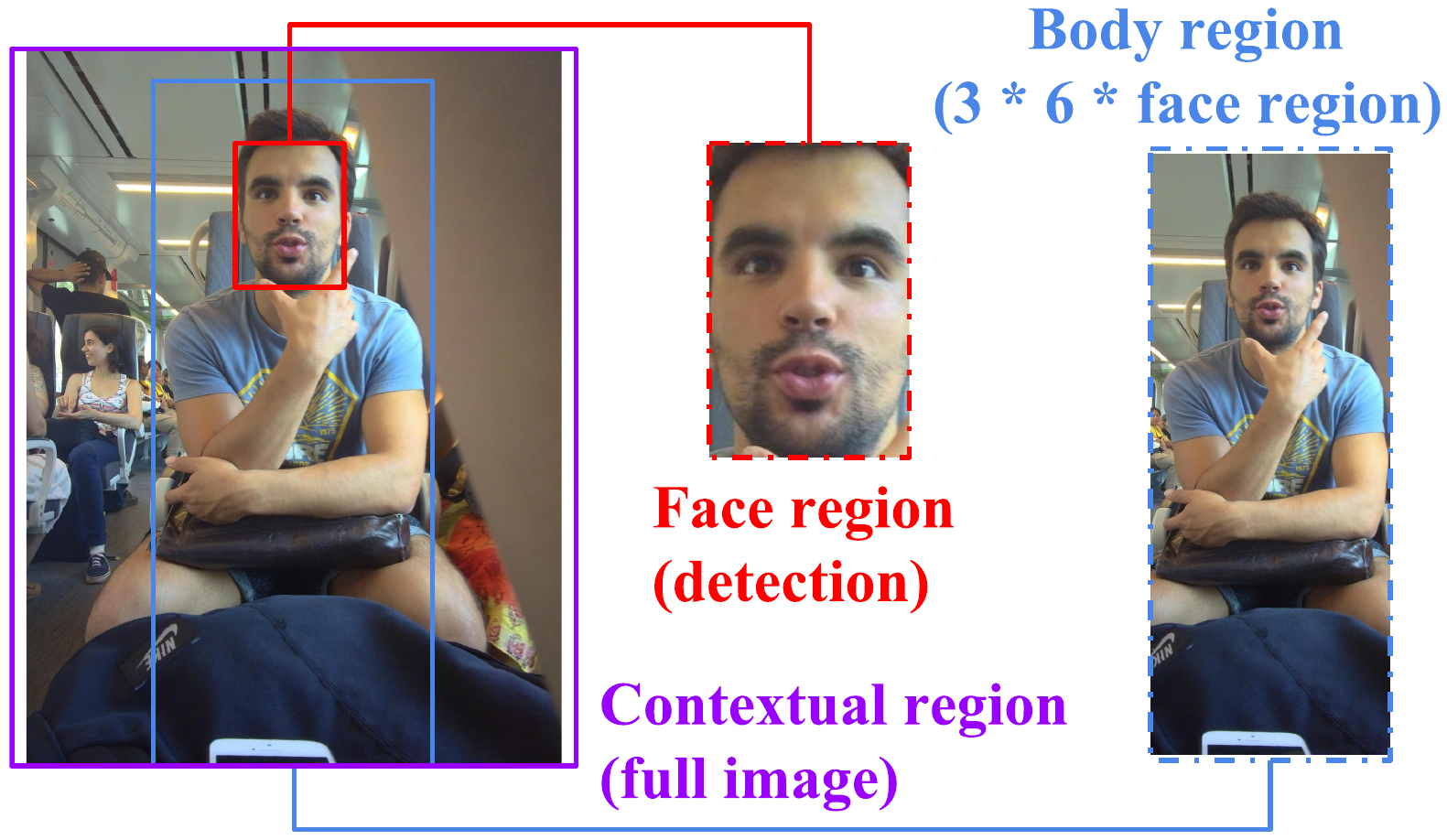}
    \caption{Illustration of face, body and contextual regions.}
    \label{fig:image_regions}
\end{figure}

In this paper, we propose a new dataset and several benchmark methods to classify social interactions from egocentric photostreams into five domains and nine relations, following the conceptualization of Bugental's theory. With this work we go beyond the state of the art on the classification of social interactions from egocentric photostreams, that is currently limited to \textit{formal/informal meeting} classification.

\vspace{-3mm}
\section{Methodology}
\label{sec:methodology}
We propose several deep learning architecture that leverage multiple semantic attributes and their temporal evolution over time. In particular, we aim at investigating the importance of semantic attributes for the classification performance in egocentric photostreams as well as how to take advantage of the hierarchical label space.  
\subsection{Preprocessing}
\label{ssec:user-segments}
Each photostream is partitioned into semantically meaningful segments by applying \textit{SRclustering} \cite{dimiccoli2017sr}, where segments with a high ratio of visible people relative to the number of frames are considered as \textit{social segments}. Given a frame in a \textit{social segment}, we extract three different regions, illustrated in fig. \ref{fig:image_regions}.  First, we apply a face detector \cite{2017RuizDockerface} to extract visible faces, discarding candidates with confidence score (IoU) below 0.99. Then, we create an initial estimate of face clusters based on visual similarity, using Microsoft Cognitive Service API. It follows a manual procedure based on visual inspection, to improve the quality of the clustering (recategorizing misclassified samples, adding uncategorized ones, discarding spurious samples and creating new clusters if needed). For each observable person, we reorganize sub-segments with valid faces. Given a valid frame, we extract \textit{Face} and \textit{Body regions}, the latter delimited by \textit{3 x face width} and \textit{6 x face height}, inspired in \cite{sun2017domain,joon2015person}. Finally, we denote the full (original) image as \textit{Contextual region}. The procedure results in a new dataset of sequences with trackable people hereafter referred to as \textit{user-specific segments}.

\subsection{Feature extraction}
\label{ssec:feat-extraction}

We leverage CNN models pretrained on specialized datasets to predict human-related attributes. Given a \textit{user-specific segment}, for each frame and social cue, we extract high-dimensional intermediate CNN features, aka \textit{visual embeddings}. We remove the task-specific classification layer, ultimately using the penultimate fully connected (FC) layer.

Table \ref{table:attributes_description_table} lists the semantic attributes used in this work. In addition, because it is not possible to observe the person holding the camera, we include the camera-wearer's ground truth age and gender information, following the categorization in \cite{sun2017domain}. If we were to concatenate all extracted features, the global representation would add up $33801$ variables. To mitigate the \textit{curse of dimensionality} phenomenon, we apply dimensionality reduction to CNN features for each attribute independently, based on the approach proposed by \cite{aghaei2018towards}. In this work, we define the quantification factor $Q = 32$, while keeping the $50$ most relevant principal components (ensuring enough level of detail, with explained variance around $90\%$). 
After merging all the semantic attributes, including compressed CNN features along with no-CNN features, we obtain a final representation with $459$ variables.

\begin{table}[]
\centering
\resizebox{0.35\textwidth}{!}{ 
\begin{tabular}{c|c|c|c|c|c}
\begin{tabular}[c]{@{}c@{}}\textbf{Semantic} \\ \textbf{attribute}\end{tabular} & \begin{tabular}[c]{@{}c@{}}\textbf{CNN} \\ \textbf{architecture}\end{tabular} & \begin{tabular}[c]{@{}c@{}}\textbf{Output} \\ \textbf{layer}\end{tabular} & \begin{tabular}[c]{@{}c@{}}\textbf{Image} \\ \textbf{region}\end{tabular} & \textbf{Dataset} & \textbf{Source}                                                 \\ \hline
\begin{tabular}[c]{@{}c@{}} (Daily)\\ Activities \end{tabular} & ResNet50 \cite{he2016ResNet}       & \textit{relu5c} & Full                                             & \cite{Cartas2018}                                                                                  & \cite{Cartas2018}                     \\ \hline
Age (x2)                                                      & CaffeNet \cite{Jia2014CaffeNet}     & \textit{relu7} & \begin{tabular}[c]{@{}c@{}} Face \\ Body \end{tabular} & PIPA \cite{joon2015person}                                                                           & \cite{sun2017domain}                        \\ \hline 
Clothing                                                      & CaffeNet \cite{Jia2014CaffeNet}     & \textit{relu7} & Body                                                    & \begin{tabular}[c]{@{}c@{}}Berkeley \\ People \\ Attributes \cite{bourdev2011describing}\end{tabular} & \cite{sun2017domain}                        \\ \hline
\begin{tabular}[c]{@{}c@{}}Facial \\ Expression\end{tabular}  & CaffeNet \cite{Jia2014CaffeNet}     & \textit{relu7} & Face                                                    & IMFDB \cite{setty2013IMFDB}                                                                        & \cite{sun2017domain}                        \\ \hline
Gender (x2)                                                   & CaffeNet \cite{Jia2014CaffeNet}     & \textit{relu7} & \begin{tabular}[c]{@{}c@{}} Face \\ Body \end{tabular} & PIPA \cite{joon2015person}                                                                           & \cite{sun2017domain}                        \\ \hline
\begin{tabular}[c]{@{}c@{}}Head\\ Appearance\end{tabular}     & CaffeNet \cite{Jia2014CaffeNet}     & \textit{relu7} & Face                                                    & CelebA \cite{Li2015}                                                                               & \cite{sun2017domain}                        \\ \hline
\begin{tabular}[c]{@{}c@{}}  Head \\ Orientation \end{tabular} & CaffeNet \cite{Jia2014CaffeNet}     & \textit{relu7} & Face                                                    & IMFDB \cite{setty2013IMFDB}                                                                        & \cite{sun2017domain}                        \\ \hline
Proximity                                                     & N/A                          & N/A & Full                                             & \cite{aghaei2018towards}                                                              & \cite{aghaei2018towards}
\end{tabular}
}

\caption{List of semantic attributes.}
\label{table:attributes_description_table}
\end{table}
\vspace{-2mm}
\subsection{Time-series classification}
\label{ssec:classification}

\begin{figure}[]
    \centering
    \includegraphics[width=0.35\textwidth]{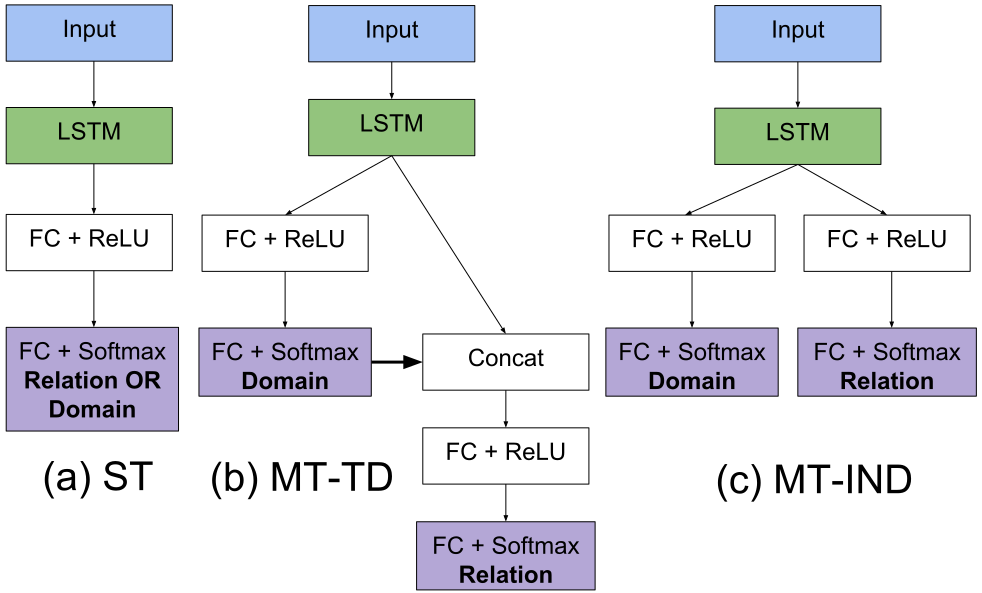}
    \caption{Evaluated architectures for social recognition.}
    \label{fig:model_archs}
\end{figure}

Given the compact representation computed for each frame in a \textit{user-specific segment}, we pose the problem of social relation recognition as multi-class time-series classification, where each component along the time axis represents the time-evolution of an individual feature. We employ a LSTM \cite{Hochreiter1997lstm}, specially designed to learn long-term dependencies in the sequence. 
Our simplest model architecture is presented in fig. \ref{fig:model_archs}a (denoted as \textit{Single-task} strategy (\textbf{ST})). Two different models are trained using either relation or domain labels. To take advantage of the hierarchical label space, we propose the model in fig. \ref{fig:model_archs}b. The class output of the first level, i.e. domains, is provided as input for predicting the second level, i.e. relations (denoted as \textit{Multi-task Top-down strategy} (\textbf{MT-TD})), inspired on the hierarchical approach proposed by Cerri \textit{et al.} \cite{cerri2016hierarchical}. We also evaluate the model in fig. \ref{fig:model_archs}c without the extra constrain (denoted as the \textit{Multi-task Independent strategy}, (\textbf{MT-IND})). As the model's description suggests, multiple objectives are trained with multi-task learning \cite{ruder2017multi_task}, jointly optimizing the loss functions (with equal importance). In all cases, the first and last FC layers are followed by \textit{ReLU} and \textit{Softmax} activation functions respectively, while using \textit{Cross-Entropy} as loss function.

\section{Experimental results}
\label{sec:results}

\subsection{Experimental setting}
\label{sec:experimental-setting}

\noindent{\bf Dataset \quad} We started from the \textit{EgoSocialStyle} dataset, collected by 9 users wearing a Narrative Clip camera, recording at two fpm in a daily life scenario. Following the protocol in \cite{aghaei2018towards}, we extended it with 119 new sequences, collected by the same users. Later on, we extracted \textit{user-specific segments} (section \ref{ssec:user-segments}) and annotated them with social labels. Similar to \cite{sun2017domain}, valid sequences must have \textbf{one valid relation label}. We discarded examples with zero or multiple labels, also with insufficient samples (less than 20). 
Our final \textit{EgoSocialRelation} dataset includes 693 sequences, grouped in five domains and nine relations, namely \textit{Attachment} (\textit{father-child}, \textit{mother-child}), \textit{Reciprocity} (\textit{friends}, \textit{classmates}), \textit{Mating} (\textit{lovers}), \textit{Coalitional group} (\textit{colleagues}), and \textit{Hierarchical group} (\textit{presenter-audience}, \textit{leader-subordinate} and \textit{customer-staff}).

\noindent{\bf Validation methodology\quad}
\label{ssec:validation} To validate our approach, we used a form of \textit{repeated random sub-sampling cross-validation} \cite{kohavi1995cross_validation}. First, we arrange our dataset in groups of whole days captured by a given user. Then, we sampled randomly $N=1000$ examples, ensuring the day's separation criteria and approximately $80\% / 20\%$ size ratios for training and validation, respectively. For each combination, we considered the best candidates with minimal  \textit{Kullback-Leibler divergence} between the normalized distributions of each split. We pick the top candidate, leaving the validation split for testing purposes, and repeat the \textit{adhoc} procedure using the training split, to obtain the top \textit{K=3} splits for model cross-validation. This strategy allows us to define data splits in a way that overlapping or consecutive user-specific sequences, that may capture the same social interaction, are put together, while maintaining the statistical distribution of the data.
Given the relatively small size of our dataset, we applied the data augmentation strategy proposed by \cite{aghaei2018towards} to mitigate the overfitting problem. In a glance, we compute PCA and add random noise in the direction of the \textit{eigenvectors}, and proportional to the \textit{eigenvalues} times a Gaussian random variable $X \sim \mathcal{N}(\mu=0,\,\sigma=0.01)$. This way we ensure that the original labels are preserved in new augmented samples. Since the dataset is highly imbalanced, we assess the competing models with two metrics, overall accuracy (abbreviated \textit{acc}) and \textit{macro f1-score}. We maximize \textit{f1-score} for model selection, giving the same importance to all classes, instead of performing well just on over-represented classes. We address class imbalance further by using a class weighting scheme embedded in the global loss function \cite{king2001_class_weights}.
It follows our final model configuration, obtained with grid search over the next parameters: number of neurons $=128$, learning rate $\alpha=2\mathrm{e}{-3}$, dropout rate $=0.3$, L2 regularization $\lambda=1\mathrm{e}{-3}$ and number of training iterations $=150$. We used Adam \cite{kingma2014adam} to optimize the model, with a \textit{step decay schedule} halving the initial $\alpha$ every 50 iterations.
\subsection{Discussion}
\label{ssec:baseline}

\noindent{\bf Social relations\quad} Although, \textit{f1-score} and \textit{acc} validate the results in distinct ways, both follow the same trend for models categorizing relations in table \ref{table:relation_results_table} (prefix \textbf{REL} and \textbf{DOM} for relation and domain recognition, respectively). By leveraging the hierarchy of labels and injecting knowledge of domains, model \textbf{REL-MT-TD} achieves the highest performance in relation recognition, this way proving beneficial to have both coarse and fine-grained social categorizations. This extra hint is key to our approach, as training with independent objectives (\textbf{REL-MT-IND}) seems counterproductive, even compared to not exploiting domain labels at all (\textbf{REL-ST}). For comparative purposes, we observe that Sun \textit{et al.} \cite{sun2017domain}, reported an \textit{acc} equivalent to \textbf{REL-ST} for relation recognition on PIPA (\textbf{REL-SVM-PIPA}, \textit{f1-score} not available).

\begin{table}[]
    \centering
    
    \resizebox{0.30\textwidth}{!}{ 
    \begin{tabular}{ccc}
    \hline
              & F1-score [\%]        & Acc  [\%]       \\ \hline
    REL-ONLY  & 32.19          & 57.10          \\ \hline
    REL-MT-I  & 31.06          & 54.90          \\ \hline
    REL-MT-TD & \textbf{33.26} & \textbf{58.60} \\ \toprule[1pt]
    DOM-ONLY          & \textbf{44.52} & \textbf{59.40} \\ \hline
    DOM-MT-I          & 38.38          & 54.90          \\ \hline
    DOM-MT-TD         & 42.49          & 56.40          \\ \toprule[1pt]
    REL-SVM-PIPA & \textbf{-} & 57.20 \\ \hline
    DOM-SVM-PIPA & \textbf{-} & 67.80 \\ \hline
    
    \end{tabular}
    }

    \caption{Social relation and domain recognition results.}
    \label{table:relation_results_table}
\end{table}

\begin{table}[]
\centering

\resizebox{0.30\textwidth}{!}{ 
\begin{tabular}{ccc}
\hline
            & \multicolumn{1}{l|}{F1-score [\%]} & \multicolumn{1}{c}{Acc [\%]} \\ \hline
REL-FACE & 23.39                       & 31.60           \\ \hline
REL-BODY & \textbf{25.30}                       & \textbf{49.60}           \\ \hline
REL-CTX  & 25.18                       & 46.60           \\ \hline
REL-ALL  & \textbf{33.26}                       & \textbf{58.60}           \\ \toprule[1pt]
DOM-FACE & \textbf{34.42}                       & 38.30           \\ \hline
DOM-BODY & 31.69                       & \textbf{50.40}           \\ \hline
DOM-CTX  & 33.61                       & 45.90           \\ \hline
DOM-ALL  & \textbf{42.49}                       & \textbf{56.40}           \\ \hline
\end{tabular}
}

\caption{Recognition results by attribute groups with MT-TD.}
\label{table:attributes_results_table}
\end{table}

\begin{figure}[t]
    \centering
        \includegraphics[width=0.3\textwidth]{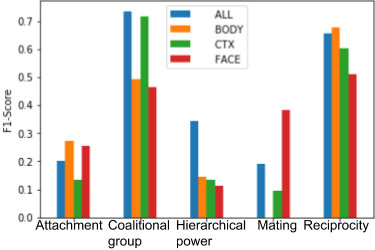}
    \caption{F1-score per domain class by group of attributes.}
    \label{fig:f1score_per_domain_class}
\end{figure}

\noindent{\bf Social domains\quad} Social characterization at domain level provides useful information, despite the coarser representation. Table \ref{table:relation_results_table} presents single-task model \textbf{DOM-ST} as the top performer for this task, surpassing alternatives that rely on relation labels, what indicates that a finer class granularity does not necessarily improve predictions at the top level. Still, model \textbf{DOM-MT-IND} has a \textit{f1-score} 10\% lower that \textbf{DOM-MT-TD}, further supporting the top-down approach. Given the hierarchical label space, we can either predict the most likely domain as presented before, or indirectly predicting the most likely relation, and then \textit{inferring} the associated domain. Notoriously, the performance of multi-task strategies increases with the second approach, giving model \textbf{DOM-MT-TD} a boost in \textit{f1-score} and \textit{acc} up to $42.69\%$ and $59.40\%$ respectively, matching the best model's \textit{accuracy}. In comparison, Sun \textit{et al.} \cite{sun2017domain} reports a slightly higher \textit{acc} of $67.80\%$ in PIPA, possibly due to a difference in criteria for model selection (we used \textit{f1-score} instead of \textit{acc}).

\noindent{\bf Analysis of semantic attributes\quad} 
In this section, we study the contribution of different subsets of semantic attributes (see table \ref{table:attributes_results_table}), with focus on the \textbf{MT-TD} strategy to simplify the analysis. Models \textbf{FACE} and \textbf{BODY} include facial and body attributes (respectively), and extra camera-user's info. Model \textbf{CTX} exploits the context by considering activity and proximity, while model \textbf{ALL} denotes the fusion of all attributes. We observe that the contribution of partial subsets of attributes is stronger for domain recognition, or equivalently, more attributes are needed to recognize relations. This support the hypothesis considering the relation recognizing as a more complex, most likely due to finer class granularity. Nevertheless, both tasks maximize \textit{f1-score} by leveraging all attributes. It can be seen that \textbf{BODY} models present considerable higher \textit{acc} than their \textbf{FACE} counterparts, most likely due to body features being more robust to partial occlusion and different perspectives. However, this fact contrasts with \textit{f1-score} performance. To shed light on this issue, fig. \ref{fig:f1score_per_domain_class} presents \textit{f1-score} computed for each domain class.

Facial attributes are specially relevant for \textit{Attachment} and \textit{Mating} (emotions, head orientation, gender cue). Furthermore, without facial information, \textit{f1-score} drops heavily for \textit{Mating}, acknowledging that lovers may not be distinguished from friends or co-workers in this scenario. In line with previous studies \cite{aghaei2018towards}, faces are key to categorize \textit{colleagues} (\textit{Coalitional group}) and \textit{friends} (\textit{Reciprocity}). Body attributes provide a different perspective, still, they are very relevant for \textit{Coalitional group} (e.g. uniform clothing) and \textit{Reciprocity}, and also for \textit{Attachment}, characterized by large age difference. Finally, daily activities (main contextual signal) are key to classify \textit{Coalitional group} (working) and \textit{Reciprocity} (gathering and sharing). Summarizing, in our experiments we observe that social domains respond to specific social cues, in correspondence with the principles proposed by Bugental.

\section{Conclusions}
\label{sec:conclusions}

This paper addressed for the first time the categorization of social relations following Bugental's conceptualization in the  domain of egocentric photostreams. A new egocentric dataset of social events acquired under unconstrained conditions, has been released and a family of models employing CNN models for feature extraction and a LSTM-based classifier have been tested providing a benchmark. Moreover, by applying multi-task learning with a hierarchical label space in a top-down approach, our model provides a solid baseline for the task of relation recognition, while outperforming straightforward alternatives. 

\clearpage

\bibliographystyle{IEEEbib}
\bibliography{mibib}

\begin{thebibliography}{10}

\bibitem{umberson2010health}
Debra Umberson and Jennifer Karas~Montez,
\newblock ``Social relationships and health: A flashpoint for health policy,''
\newblock {\em Journal of health and social behavior}, vol. 51, no. 1\_suppl,
  pp. S54--S66, 2010.

\bibitem{Ramanathan2013}
V.~Ramanathan, B.~Yao, and L.~Fei-Fei,
\newblock ``Social role discovery in human events,''
\newblock in {\em IEEE CVPR}, 2013, pp. 2475--2482.

\bibitem{dehghan2014look}
Afshin Dehghan, Enrique~G Ortiz, Ruben Villegas, and Mubarak Shah,
\newblock ``Who do i look like? determining parent-offspring resemblance via
  gated autoencoders,''
\newblock in {\em IEEE CVPR}, 2014, pp. 1757--1764.

\bibitem{murillo2012urban}
Ana~C Murillo, Iljung~S Kwak, Lubomir Bourdev, David Kriegman, and Serge
  Belongie,
\newblock ``Urban tribes: Analyzing group photos from a social perspective,''
\newblock in {\em CVPRW}. IEEE, 2012, pp. 28--35.

\bibitem{shao2013you}
Ming Shao, Liangyue Li, and Yun Fu,
\newblock ``What do you do? occupation recognition in a photo via social
  context,''
\newblock in {\em ICCV IEEE}, 2013, pp. 3631--3638.

\bibitem{sun2017domain}
Qianru Sun, Bernt Schiele, and Mario Fritz,
\newblock ``A domain based approach to social relation recognition,''
\newblock in {\em IEEE CVPR}, 2017, pp. 21--26.

\bibitem{aghaei2018towards}
Maedeh Aghaei, Mariella Dimiccoli, Cristian~Canton Ferrer, and Petia Radeva,
\newblock ``Towards social pattern characterization in egocentric
  photo-streams,''
\newblock {\em Computer Vision and Image Understanding}, vol. 171, pp.
  104--117, 2018.

\bibitem{bolanos2017towards}
Marc Bolanos, Mariella Dimiccoli, and Petia Radeva,
\newblock ``Toward storytelling from visual lifelogging: An overview,''
\newblock {\em IEEE Transactions on Human-Machine Systems}, vol. 47, no. 1, pp.
  77--90, 2017.

\bibitem{Bugental2000}
Daphne~Blunt Bugental,
\newblock ``Acquisition of the algorithms of social life: a domain-based
  approach.,''
\newblock {\em Psychological bulletin}, vol. 126 2, pp. 187--219, 2000.

\bibitem{Fathi2012}
Alireza Fathi, Jessica Hodgins, and James Rehg,
\newblock ``Social interactions: A first-person perspective,''
\newblock pp. 1226--1233, 06 2012.

\bibitem{yang2016wearable}
Jen-An Yang, Chia-Han Lee, Shao-Wen Yang, V~Srinivasa Somayazulu, Yen-Kuang
  Chen, and Shao-Yi Chien,
\newblock ``Wearable social camera: Egocentric video summarization for social
  interaction,''
\newblock in {\em ICMEW IEEE}, 2016, pp. 1--6.

\bibitem{aghaei2015towards}
Maedeh Aghaei, Mariella Dimiccoli, and Petia Radeva,
\newblock ``Towards social interaction detection in egocentric photo-streams,''
\newblock in {\em ICMV}, 2015, vol. 9875.

\bibitem{aghaei2016whom}
Maedeh Aghaei, Mariella Dimiccoli, and Petia Radeva,
\newblock ``With whom do i interact? detecting social interactions in
  egocentric photo-streams,''
\newblock in {\em ICPR 2016}. IEEE, 2016, pp. 2959--2964.

\bibitem{Hochreiter1997lstm}
Sepp Hochreiter and J{\"u}rgen Schmidhuber,
\newblock ``Long short-term memory,''
\newblock {\em Neural computation}, vol. 9, no. 8, pp. 1735--1780, 1997.

\bibitem{dimiccoli2017sr}
Mariella Dimiccoli, Marc Bola{\~n}os, Estefania Talavera, Maedeh Aghaei,
  Stavri~G Nikolov, and Petia Radeva,
\newblock ``Sr-clustering: Semantic regularized clustering for egocentric photo
  streams segmentation,''
\newblock {\em Computer Vision and Image Understanding}, vol. 155, pp. 55--69,
  2017.

\bibitem{2017RuizDockerface}
N.~{Ruiz} and J.~M. {Rehg},
\newblock ``{Dockerface: an easy to install and use Faster R-CNN face detector
  in a Docker container},''
\newblock {\em ArXiv e-prints}, Aug. 2017.

\bibitem{joon2015person}
Seong Joon~Oh, Rodrigo Benenson, Mario Fritz, and Bernt Schiele,
\newblock ``Person recognition in personal photo collections,''
\newblock in {\em ICCV IEEE}, 2015, pp. 3862--3870.

\bibitem{he2016ResNet}
Kaiming He, Xiangyu Zhang, Shaoqing Ren, and Jian Sun,
\newblock ``Deep residual learning for image recognition,''
\newblock in {\em IEEE CVPR}, 2016, pp. 770--778.

\bibitem{Cartas2018}
Alejandro Cartas, Juan Mar{\'i}n, Petia Radeva, and Mariella Dimiccoli,
\newblock ``Batch-based activity recognition from egocentric photo-streams
  revisited,''
\newblock {\em Pattern Analysis and Applications}, vol. 21, no. 4, pp.
  953--965, 2018.

\bibitem{Jia2014CaffeNet}
Yangqing Jia, Evan Shelhamer, Jeff Donahue, Sergey Karayev, Jonathan Long, Ross
  Girshick, Sergio Guadarrama, and Trevor Darrell,
\newblock ``Caffe: Convolutional architecture for fast feature embedding,''
\newblock in {\em ACM Multimedia}, 2014, pp. 675--678.

\bibitem{bourdev2011describing}
Lubomir Bourdev, Subhransu Maji, and Jitendra Malik,
\newblock ``Describing people: A poselet-based approach to attribute
  classification,''
\newblock in {\em IEEE ICCV}, 2012, pp. 1543--1550.

\bibitem{setty2013IMFDB}
Shankar Setty, Moula Husain, Parisa Beham, Jyothi Gudavalli, Menaka Kandasamy,
  Radhesyam Vaddi, Vidyagouri Hemadri, JC~Karure, Raja Raju, B~Rajan, et~al.,
\newblock ``Indian movie face database: a benchmark for face recognition under
  wide variations,''
\newblock in {\em CVPRIPG}. IEEE, 2013, pp. 1--5.

\bibitem{Li2015}
Li-Jia Li, David~A. Shamma, Xiangnan Kong, Sina Jafarpour, Roelof van Zwol, and
  Xuanhui Wang,
\newblock ``Celebritynet: A social network constructed from large-scale online
  celebrity images,''
\newblock {\em TOMCCAP}, vol. 12, pp. 3:1--3:22, 2015.

\bibitem{cerri2016hierarchical}
Ricardo Cerri, Rodrigo~C Barros, Andr{\'e}~CPLF de~Carvalho, and Yaochu Jin,
\newblock ``Reduction strategies for hierarchical multi-label classification in
  protein function prediction,''
\newblock {\em BMC bioinformatics}, vol. 17(1), pp. 373, 2016.

\bibitem{ruder2017multi_task}
Sebastian Ruder,
\newblock ``An overview of multi-task learning in deep neural networks.,''
\newblock {\em CoRR}, vol. abs/1706.05098, 2017.

\bibitem{kohavi1995cross_validation}
Ron Kohavi,
\newblock ``A study of cross-validation and bootstrap for accuracy estimation
  and model selection,''
\newblock in {\em IJCAI - Volume 2}, 1995, pp. 1137--1143.

\bibitem{king2001_class_weights}
Gary King and Langche Zeng,
\newblock ``Logistic regression in rare events data,''
\newblock {\em Political analysis}, vol. 9, no. 2, pp. 137--163, 2001.

\bibitem{kingma2014adam}
Diederik~P Kingma and Jimmy Ba,
\newblock ``Adam: A method for stochastic optimization,''
\newblock {\em arXiv preprint arXiv:1412.6980}, 2014.

\end{thebibliography}

\end{document}